\documentclass[10pt,twocolumn,letterpaper]{article}

\usepackage{wacv}
\usepackage{times}
\usepackage{epsfig}

\usepackage{enumitem}
\usepackage{graphicx}
\graphicspath{{images/}}
\usepackage{epstopdf}
\DeclareGraphicsExtensions{.pdf,.jpeg,.png,.eps}

\usepackage{subfig}
\usepackage{array}
\usepackage{mdwmath}
\usepackage{mdwtab}
\usepackage{eqparbox}

\usepackage[pagebackref=true,breaklinks=true,letterpaper=true,colorlinks,bookmarks=false]{hyperref}

\wacvfinalcopy 

\def\wacvPaperID{****} 

\ifwacvfinal\fi
\begin{document}

\title{Energy-Efficient ConvNets Through Approximate Computing}

\author{Bert Moons$^\dagger$\thanks{Equally contributing authors.} \hspace{1cm} Bert De Brabandere$^\circ$$^\ast$ \hspace{1cm} Luc Van Gool$^\circ$ \hspace{1cm} Marian Verhelst$^\dagger$ \\
$^\dagger$ESAT-MICAS, $^\circ$ESAT-VISICS, Dept. of Electrical Engineering, KU Leuven\\
{\tt\small bert.moons@esat.kuleuven.be, bert.debrabandere@esat.kuleuven.be}
}

\maketitle
\ifwacvfinal\thispagestyle{empty}\fi

\begin{abstract}
   Recently ConvNets or convolutional neural networks (CNN) have come up as state-of-the-art classification and detection algorithms, achieving near-human performance in visual detection. However, ConvNet algorithms are typically very computation and memory intensive. In order to be able to embed ConvNet-based classification into wearable platforms and embedded systems such as smartphones or ubiquitous electronics for the internet-of-things, their energy consumption should be reduced drastically. This paper proposes methods based on approximate computing to reduce energy consumption in state-of-the-art ConvNet accelerators. By combining techniques both at the system- and circuit level, we can gain energy in the systems arithmetic: up to $30\times$ without losing classification accuracy and more than $100\times$ at $99\%$ classification accuracy, compared to the commonly used 16-bit fixed point number format.
\end{abstract}

\section{Introduction}

Recently neural networks have made an impressive comeback in the field of machine learning. ConvNets or Convolutional neural networks (CNN) are consistently pushing the state-of-the-art in areas like computer vision and speech processing. One of the reasons for this revival is the increasing availability of computing power. Multicore CPU's, GPU's, and even clusters of GPU's are no longer prohibitively expensive and make it possible to train and evaluate larger networks.

Unfortunately, the increase in computing power is inevitably accompanied by an increase in energy consumption. While high energy consumption is no big concern during the network's training phase - which typically takes place on a computer cluster - it poses a problem when the network needs to be evaluated on mobile hardware like smartphones, smart glasses and other wearable devices.

In this paper, we investigate how energy consumption can be reduced through principles of approximate computing. Our analysis shows that it is possible to quantize existing networks and drastically reduce the number of bits that encode the weights and inputs of each layer, with only a minimal loss in accuracy and without the need to retrain the networks. With an appropriate hardware architecture this knowledge can be used to reduce energy consumption in many neural network applications and even dynamically scale energy consumption depending on the current application.

Our contribution is threefold:
\begin{enumerate}
	\item We show how computational precision can be scaled in several ConvNet architectures for image classification through quantization of the layer's inputs and weights.  The possible quantization varies per architecture, per application and even per layer within a single ConvNet.
	\item We show how this approximate computing / precision scaling can lead to reduced energy consumption in ConvNet-accelerators. This is achieved through a combination of algorithmical and circuit-level techniques:

	\textbf{-} Applying precision scaling on ConvNet-accelerator circuits.

	\textbf{-} The number of zero-valued parameter and input-values of ConvNet-layers increases through precision scaling. Their computations can be skipped. 
 
	\item Based on this analysis of algorithmic accuracy and energy savings, we demonstrate achievable energy-accuracy curves for three popular ConvNet architectures for image classification.
\end{enumerate}

\section{Related work}

While ConvNets have a long history in the field of computer vision, only recently they have become the go-to technique for classification and detection problems. This newfound popularity is a result of their impressive performance on a number of benchmarks, reaching super-human accuracy on tasks like handwritten character recognition \cite{lecun1998mnist} and near-human performance on large scale classification challenges like ILSVRC \cite{deng_imagenet:_2009}. Record-breaking ConvNet architectures are being developed in academia \cite{krizhevsky_imagenet_2012}~\cite{Simonyan14c} and industry \cite{szegedy_going_2014}~\cite{wu_deep_2015}.

A ConvNet has a hierarchical structure consisting of a number of layers. Most architectures contain multiple blocks of convolutional layers, rectified linear units (ReLU) and pooling layers, followed by a number of fully connected layers. The network is trained with the backpropagation algorithm, which iteratively updates the weights in the convolutional and fully connected layers in order to minimize a certain loss function. 

The success of these algorithms has led to the interest of the embedded vision community. A number of ConvNet hardware optimizations \cite{conti2015ultra} \cite{peemen2013memory}, coprocessors \cite{farabet2011neuflow} and accelerators such as DianNao \cite{chen2014diannao} have been proposed. Although all of these works meant a significant step forward, the realization of a high-performance, energy-efficient and real-time ConvNet-architecture, which could work autonomously from the cloud is not yet complete. The optimized architectures named above, can however be optimized further by exploiting ConvNet's inherent error-resilience.  The effect of quantization errors on regular neural networks have been studied in \cite{jiang_effects_2003} and \cite{dundar1994effects}.

Error resilient applications can be made more energy efficient through approximate computing. This is a collection of hardware-level techniques enabling energy savings at the expense of reduced computational accuracy. This can be achieved in various ways, leading to either a static (fixed after design-time) or a dynamic (adaptable after design-time) trade-off. Static examples in literature make use of approximate digital building blocks for arithmetic functions~\cite{kulkarni2011trading}, or use approximate logic synthesis methods~\cite{venkataramani2012salsa}. Dynamic approximate computing mainly uses the principle of precision scaling, in which the number of bits used in computations is varied at run-time 
\cite{venkataramani2013quality}~\cite{de2012flexible}~\cite{moons2015DVAS}. In this work, we use precision scaling to achieve energy gains in ConvNet-acceleration. Since precision scaling can be applied dynamically, the energy consumption of the targeted accelerator can be adapted to the varying precision requirements of the used ConvNet network and even of the ConvNet network layer.

\begin{figure}[!t]
	\centering
	\subfloat[]{{\label{fig:multiplier_full}\includegraphics[width=0.25\textwidth]{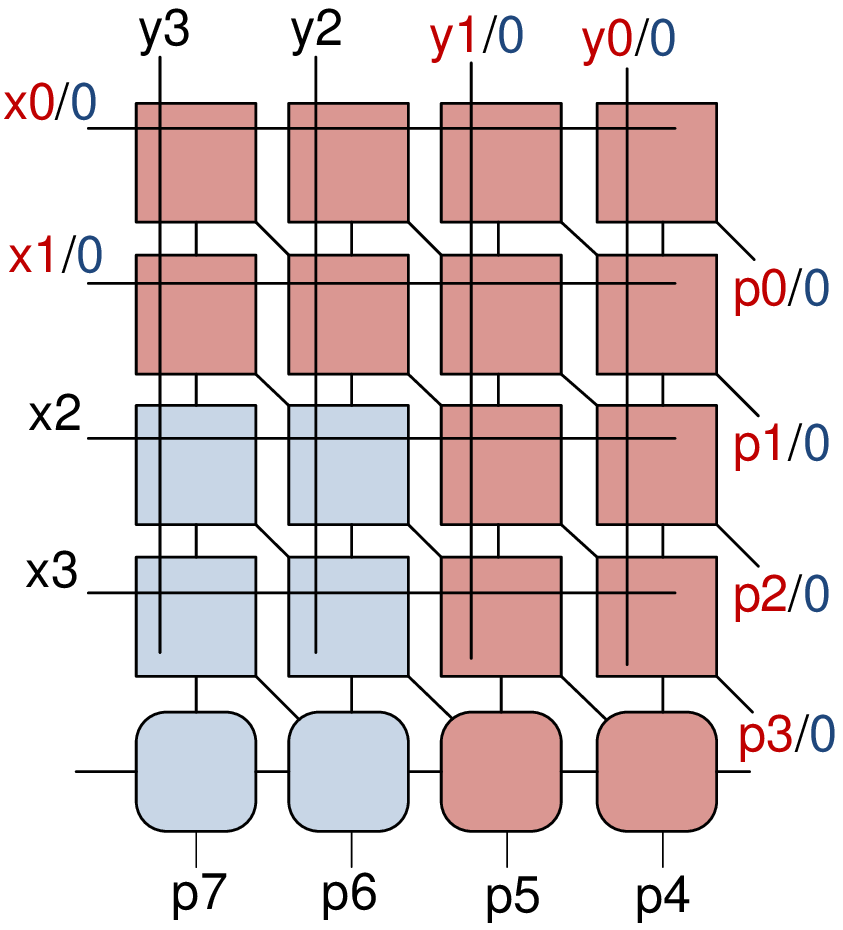} }}%
	\quad
	\subfloat[]{{\label{fig:power_vs_rmse}\includegraphics[width=0.37\textwidth]{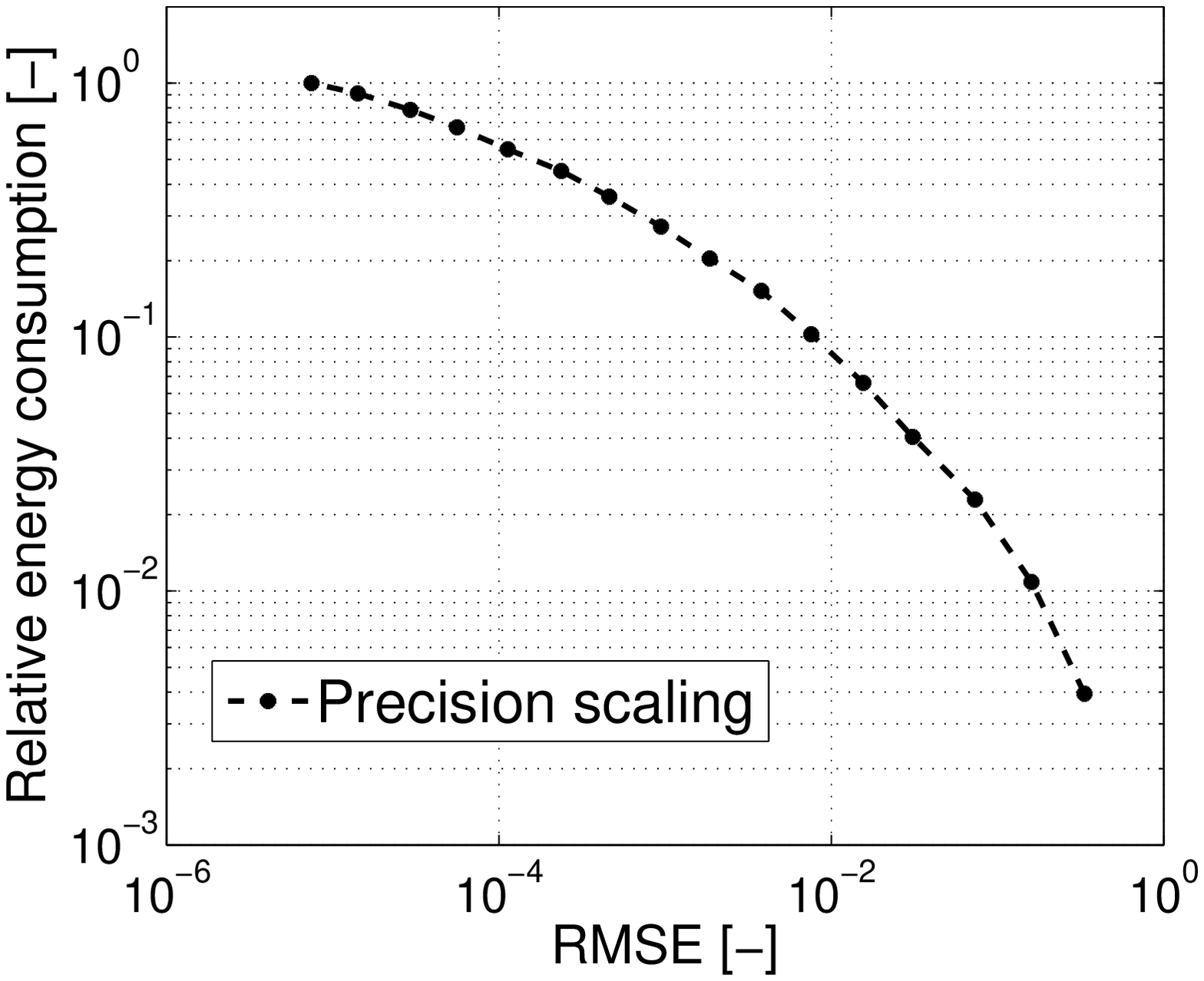} }}%
	\caption{Energy-accuracy trade-off in digital multipliers through precision scaling (a) By scaling the used number of bits from $4$ to $2$ the number of active blocks drops from $20$ to $6$. (b) Energy consumption versus root-mean-square-error under precision scaling. Precision scaling can gain $12\times$ in energy at 1\% RMSE.}%
	\label{fig:dvas}%
\end{figure}

Approximate computing through precision scaling lowers the active power consumption of a digital circuit.
The total power consumption consists of a dynamic and a leakage component. It can be summarized as P = $\alpha C f V^2 + P_{leakage}$
where $\alpha$ is the circuit switching activity, $f$ is the clock frequency, $C$ is the total switching capacitance and $V$ is the circuit's supply voltage. By scaling precision (i.e., dynamically scaling the number of bits that encode the network's weights and inputs), the switching activity $\alpha$ can be substantially reduced. Figure~\ref{fig:dvas} illustrates the achievable energy-accuracy trade-off in a typical digital multiplier. Note that a striking $12\times$ energy gain can be achieved at an output error of 1\% root-mean-square-error (RMSE).

A first attempt in combining neural networks with approximate computing has been done in \cite{venkataramani2014axnn}. Their work only looks into regular neural networks and requires retraining of the network. In this work we focus on convolutional neural network architectures, which perform much better in computer vision applications. Furthermore, we do not require network retraining.

\section{Analysis and experiments}
We examine the performance- and energy-related effects of quantization on three different network architectures for image classification. Because it is impossible to cover the entire spectrum of architectures and applications, we choose three popular networks as representatives of small-, medium-, and large scale architectures that differ significantly in their number of parameters:
\begin{itemize}
	\item LeNet-5 on MNIST \cite{le1990handwritten}: LeNet-5 is a small network with two convolutional and two fully connected layers. It is designed to classify handwritten digits (20x20 images) in the MNIST dataset where it achieves an accuracy of 99.0\%.
	\item CifarQuick on Cifar-10 \cite{krizhevsky2009learning}: CifarQuick is a medium-sized architecture with three convolutional and two fully connected layers. It has more filters per layer than LeNet-5 and operates on color images instead of grayscale images. It classifies images of the Cifar-10 dataset, which consists of small 32x32 color images divided into ten classes. It reaches an accuracy of 75.3\%.
	\item AlexNet on ImageNet \cite{krizhevsky_imagenet_2012}: AlexNet is a large network with five convolutional and three fully connected layers. It reaches a top-5 accuracy of 80.0\% on ILSVRC2012, a large scale classification challenge with 1000 categories.
\end{itemize}

For our experiments we customized the open-source deep learning framework Caffe \cite{Jia2014} to be able to simulate quantization of the network's weights and inputs. All experiments are run on the validation sets of the discussed benchmarks. We always report the relative accuracy, i.e. the ratio of the accuracy after quantization and the accuracy of the original network. 

Note that our techniques can be used to reduce the energy consumption during evaluation of the network, but not necessarily during the training phase. When training the network, the weight quantization would cause the backpropagation algorithm to quickly get stuck in local optima.

\subsection{Influence of quantization on classification accuracy}
\label{subsec:InfluenceOfQuantizationOnAccuracy}
Typically, ConvNets run on high precision machines using 32-bit floating point number representations. Dedicated embedded platforms use 16-bit fixed point hardware for ConvNet computations. However, such high precision is not always necessary in. The energy spent in high precision computations, does not lead to more accurate classification by the algorithm.
To reduce the energy consumption of the ConvNet's computations, our main strategy is to quantize its weights and the inputs to its layers. Such quantization leads to a network that is only an approximation of the original network. Our goal is to find out the influence of quantization on a network's accuracy and whether the effects differ significantly across network architectures. We are interested in how far we can push the quantization of a network without sacrificing too much classification accuracy.

Before quantizing the weights and inputs, it is important to rescale them properly according to the distribution of their values. If there is a mismatch between the interval in which these values lie and the interval over which we quantize, the accuracy will drop even at very low quantization settings. For this reason, we rescale all layer inputs and weights in the network with a single value that corresponds to the maximal input or weight value (rounded to the next factor of two) observed during a complete run over the data. This ensures that the limits of the quantization interval correspond to the limits of the data, and no bits are wasted. Mathematically, this scaling is the same as multiplying the results with a factor of 2, or shifting data in a fixed point number representation. This operation has a limited hardware energy footprint and only needs to be performed at convolutional outputs.

\begin{figure}[!t]
	\centering
	\subfloat[]{{\label{fig:bitsVsAccuracy_a}\includegraphics[width=0.40\textwidth]{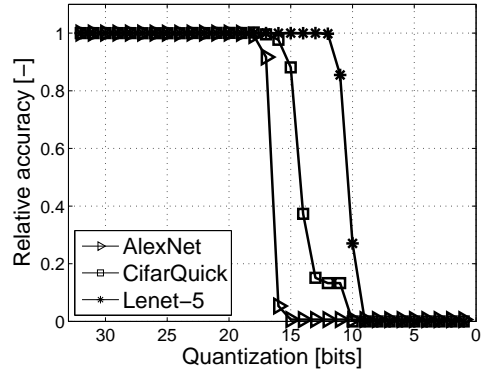} }}%
	\quad
	\subfloat[]{{\label{fig:bitsVsAccuracy_b}\includegraphics[width=0.40\textwidth]{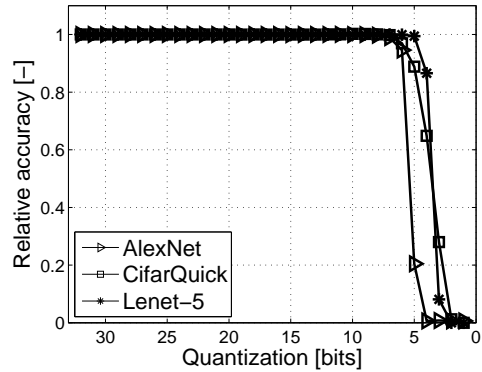} }}%
	\caption{Relative accuracy of three architectures as a function of the number of quantization bits. Layer inputs and weights are first rescaled so they can be quantized more effectively. We compare two rescaling strategies: (a) Uniform rescaling: all inputs and weights are scaled with the same value. (b) Per-layer rescaling: inputs and weights are scaled on a per-layer basis.}%
	\label{fig:bitsVsAccuracy}%
\end{figure}

\subsubsection{Uniform quantization}
As a first experiment we choose a single quantization setting for all layers in the network, and call this \emph{uniform quantization}. Later, we also introduce \emph{per-layer quantization}, where each layer is quantized separately.

Figure~\ref{fig:bitsVsAccuracy_a} shows the relative accuracy of our three networks as a function of the number of quantization bits. As can be seen in the figure, the relative accuracy stays equal to one for all three networks up to quantization with 19 bits, meaning that the quantized network reaches the exact same accuracy on its dataset as the original network. At a quantization with 18-bit however, the accuracy for AlexNet starts dropping quickly, rendering it useless for any practical application. At 11-bit the same effect can be seen for the smaller LeNet-5.

\begin{figure}[!t]
	\centering
	\subfloat[]{{\label{fig:caffeOutputStatistics_a}\includegraphics[width=0.40\textwidth]{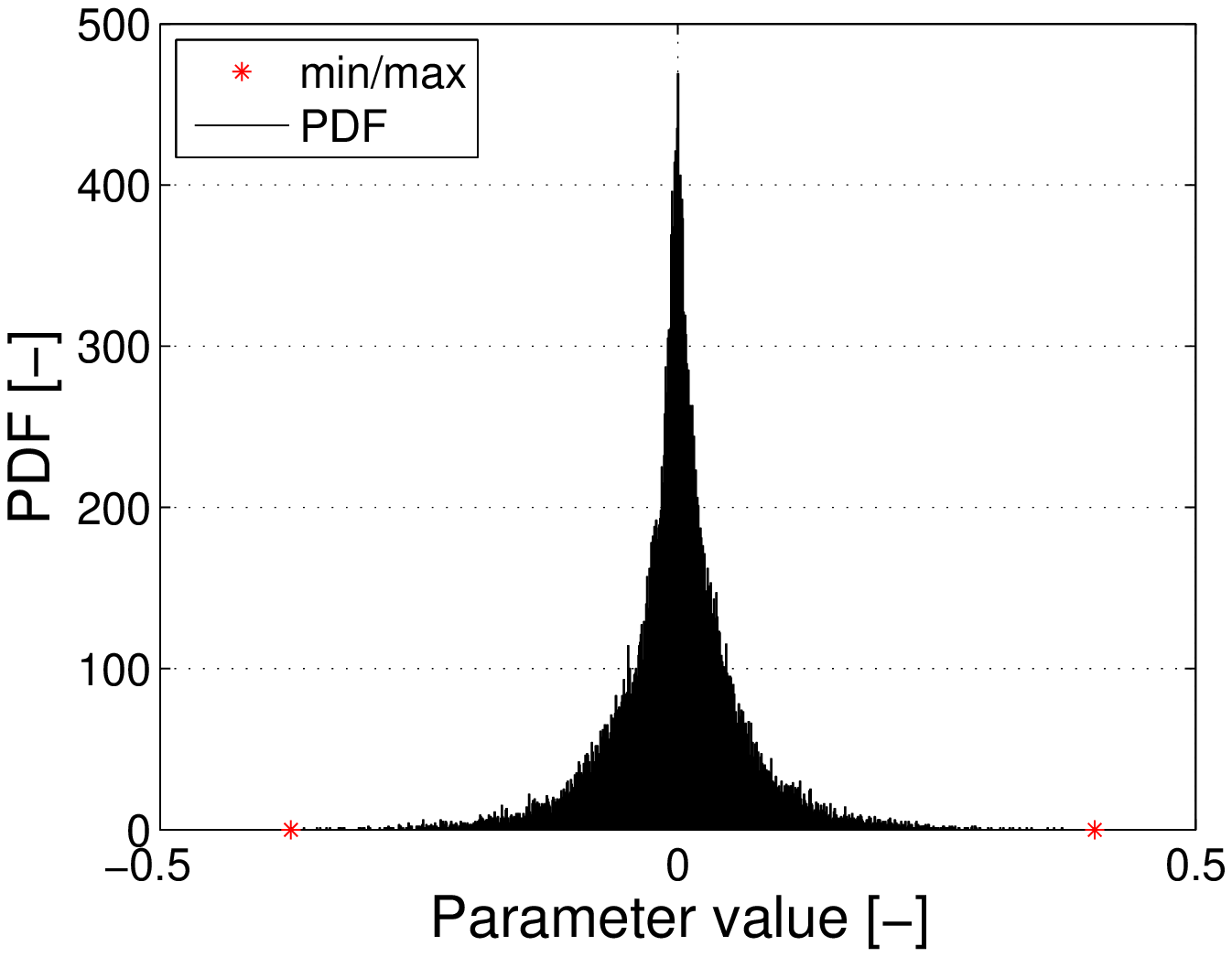} }}%
	\quad
	\subfloat[]{{\label{fig:caffeOutputStatistics_b}\includegraphics[width=0.40\textwidth]{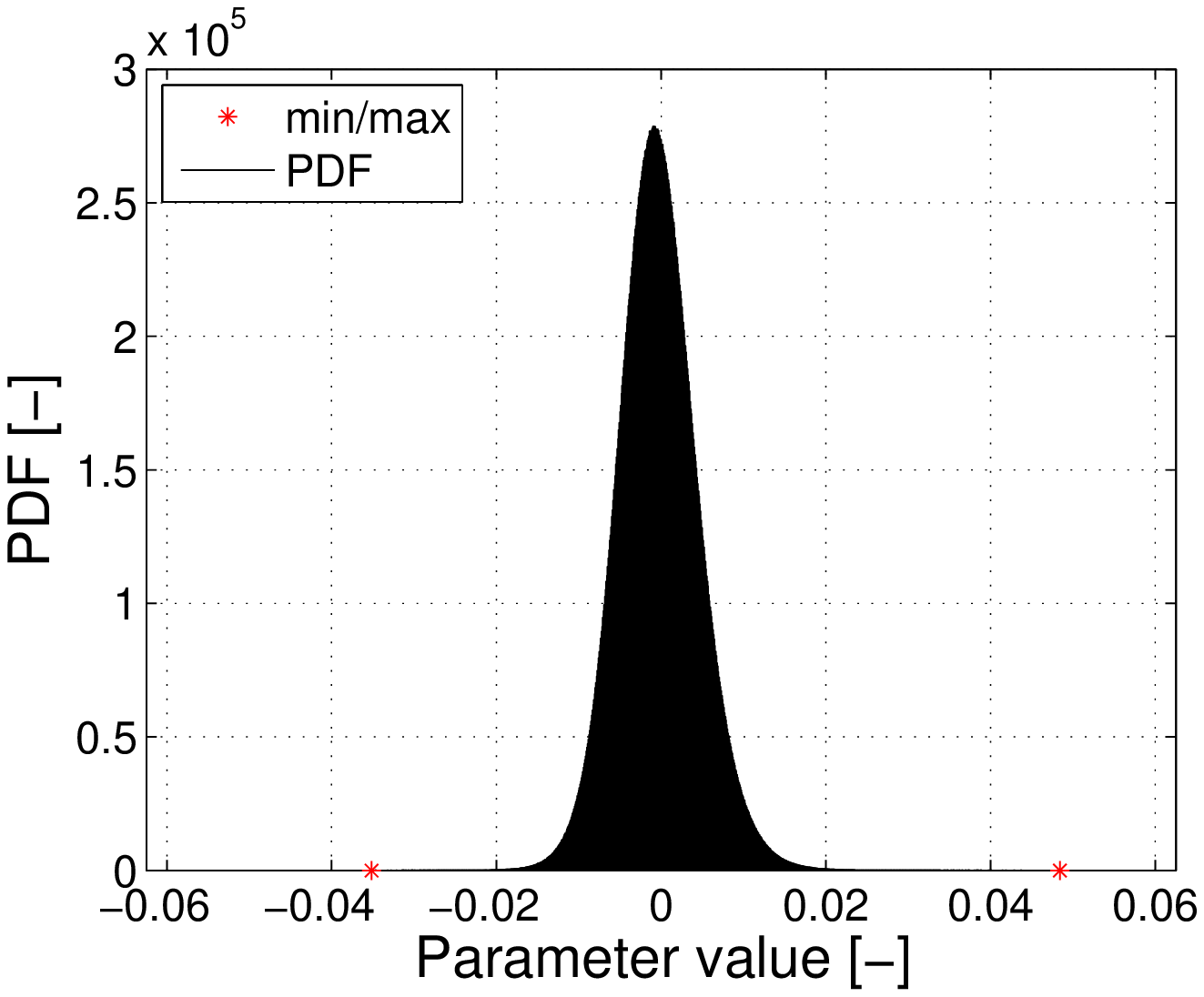} }}%
	\caption{AlexNet's first and sixth layers weight statistics. Layer 1 should be projected on the [-0.5,0.5] interval and layer 6 on the [-0.0625,0.0625] interval to minimize the needed number of bits. If layer 6 would be projected on the [-1,1] interval, 4 extra bits would be needed.}%
	\label{fig:caffeOutputStatistics}%
\end{figure}

We can do better than this by applying the scaling in a more fine-grained way; we choose a different scaling factor for each layer. This is clarified by a simple example: AlexNet's first and sixth layer weight statistics are shown in Figure~\ref{fig:caffeOutputStatistics}. All weights of layer 1 are within the $[-0.5,0.5]$ interval, while the weights of layer 6 are within the $[-0.0625, 0.0625]$ interval. By allowing to quantize the weights in layer 6 over this smaller interval instead of over $[-0.5,0.5]$ we recover 3 bits that would otherwise be wasted. A similar reasoning holds for the layer's inputs.

The effect of the per-layer scaling is significant, as can be seen in Figure~\ref{fig:bitsVsAccuracy_b}. Compared to the previous , we can now quantize much more aggressively without sacrificing accuracy. Each network performs at its original accuracy up to quantization with no more than 8 bits. After that, the accuracy again drops quickly. The reason for this improvement is the fact that input and weight statistics differ greatly among layers. Per-layer rescaling allows to set the optimal quantization interval in each layer.

\subsubsection{Per-layer quantization}
\label{subsubsec:nonuniformQuantization}
Just as we do per-layer rescaling, we can also do per-layer quantization: instead of quantizing all weights and inputs of the network with the same number of bits, we choose a different setting in each layer. The idea is again to find an optimal setting for each layer by exploiting the variations of the layer-specific input and weight distributions. Another effect of the per-layer quantization is that we can set the operating point (i.e. the desired minimal relative accuracy) more precisely and thus control the energy-accuracy trade-off tightly, as discussed later in section~\ref{subsec:energyAccuracyTradeoff}. This is not possible with uniform quantization: e.g. between quantization with 5 and 4 bits, the relative accuracy of LeNet-5 drops from 99.4\% immediately to an unusable 86.6\%.

In order to find a good quantization setting for each layer, we do a greedy search over the parameters: starting at the first layer, its input is quantized until the accuracy drops to the target accuracy. Next, the quantization of the input is kept fixed while the quantization of its weights is maximized in the same way. The same process is applied in the next layers until the last one. If a value goes out of range, we clip it to the respective minimum or maximum representable value with the chosen quantization. The full sweep can take many hours, depending on the network size. This poses no real problem, as the procedure has to be performed only once and can be done off-line.

\begin{figure*}[!t]
	\begin{center}
		\centerline{
		\subfloat[]{{\label{fig:nonUniformQuantization_a}\includegraphics[width=0.37\textwidth]{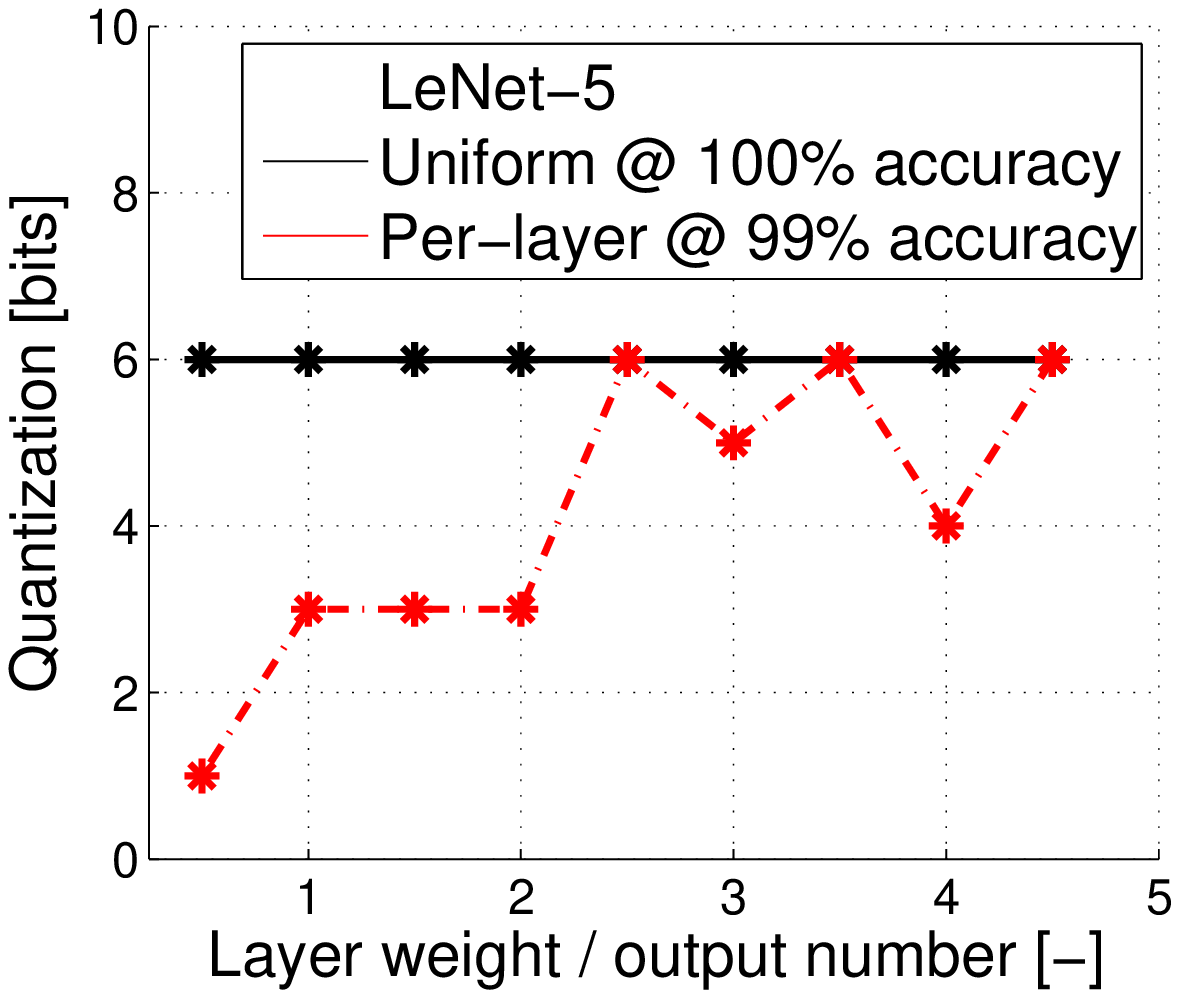} }}
		\subfloat[]{{\label{fig:nonUniformQuantization_b}\includegraphics[width=0.37\textwidth]{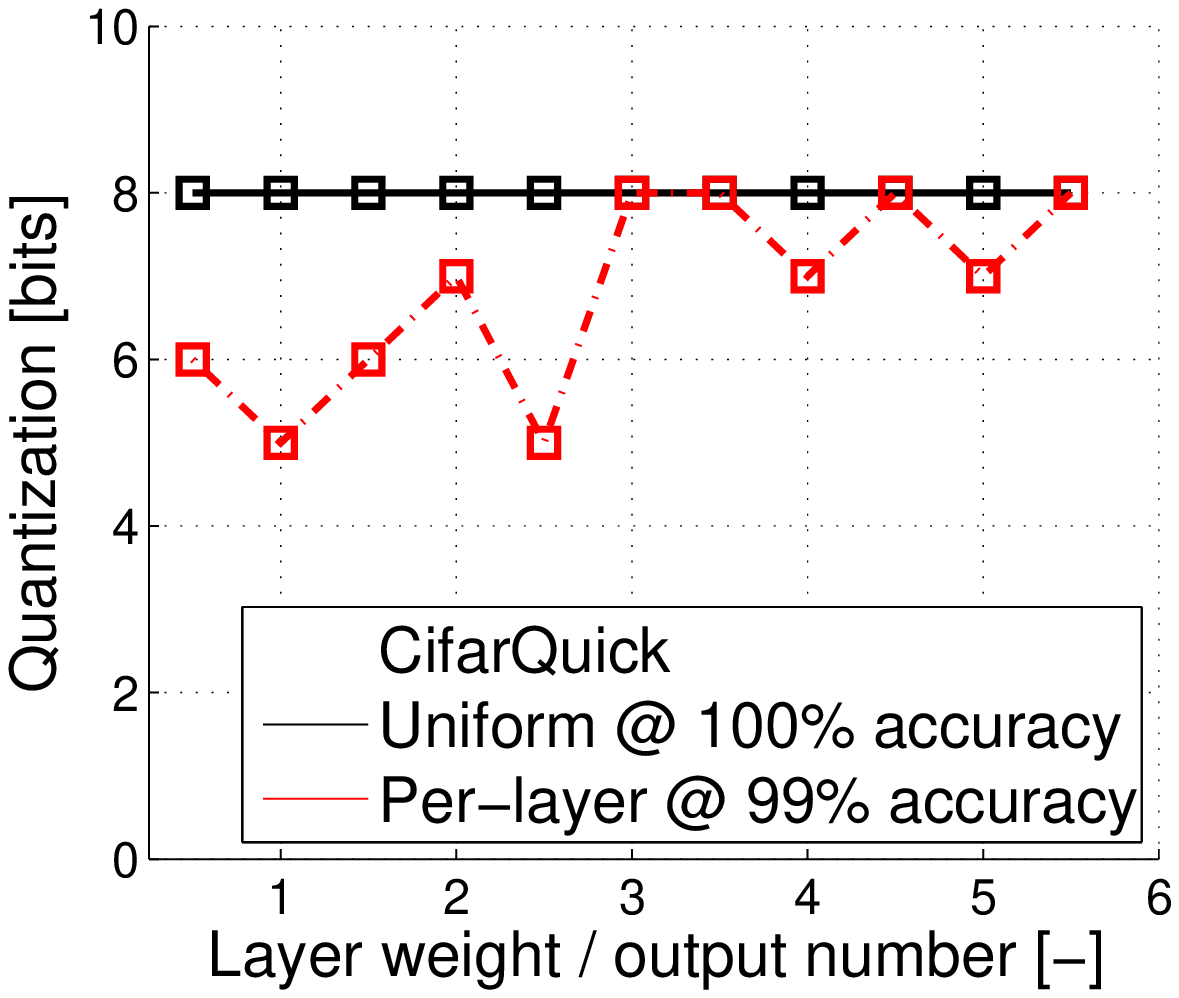} }}
		\subfloat[]{{\label{fig:nonUniformQuantization_c}\includegraphics[width=0.37\textwidth]{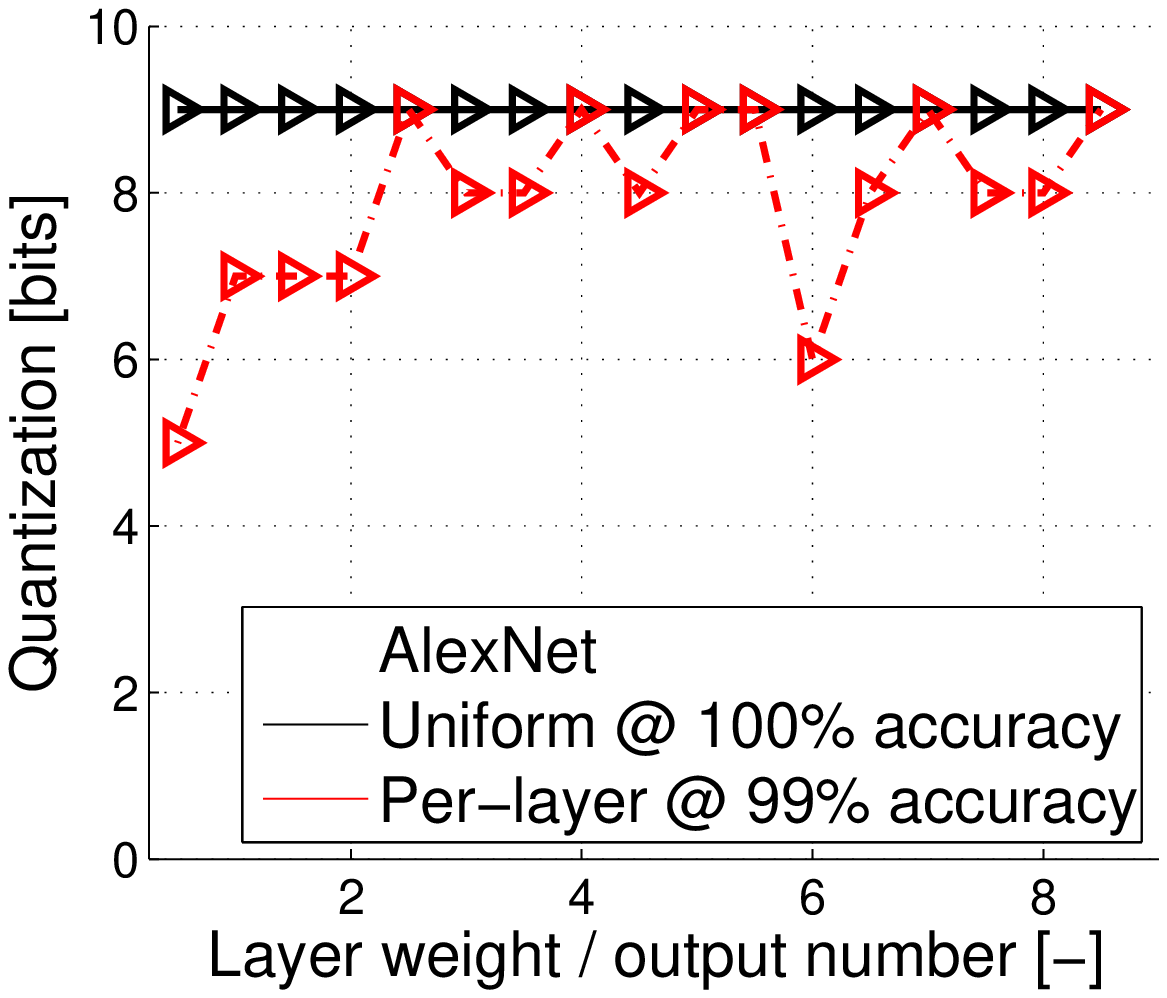} }}
		}
	\end{center}
	\caption{Comparison of uniform and optimal per-layer quantization. Data points at integer values show quantization of the layer weights. Other data points show quantization of the layer inputs. Maximum achievable quantizations: (a) LeNet-5: 1-6 bit. (b) CifarQuick: 5-8 bit. (c) AlexNet: 5-9 bit. }%
	\label{fig:nonUniformQuantization}%
\end{figure*}

The amount of bits that can be saved with per-layer quantization at a target accuracy of 99\%, compared to uniform quantization at 100\% relative accuracy, is visualized for each reference network in Figure~\ref{fig:nonUniformQuantization}. The results are ad-hoc, but there is a general trend of needing less bits in the lower layers of the network than in the higher layers. This is partly a result of the forward parameter sweep, but we hypothesize that the difference in input and weight statistics between lower and higher layers also plays a significant role.


\subsection{Proposed energy reduction through precision scaling and computation skipping}
This section indicates how this increased quantization can lead to energy savings in real hardware architectures.
To quantify the possible energy gains of approximate computing in ConvNet-acceleration, we model the energy consumption of the necessary convolutional arithmetic (multiply and add) for a full ConvNet-algorithm. This analysis does not incorporate the energy overheads of control, I/O, data- and program-memory interfacing and the clock network, as these are highly architecture dependent. It can thus be considered as the maximum potential energy savings through approximate computing. 
Furthermore, the number of zero-valued weights and inputs of typical ConvNets increases at higher quantization. This can be exploited algorithmically by skipping unnecessary computations with zero-valued inputs. This will lead to large additional energy savings.

\subsubsection{Precision scaling}
Precision scaling is an approximate computing technique allowing major energy savings in digital circuits. Lower precision computing, i.e. computing using less bits, reduces the switching activity. Figure~\ref{fig:dvas} and \cite{moons2015DVAS} show how this concept can be applied to a common digital multiplier.

\begin{table}[]
	\renewcommand{\arraystretch}{1.3}
	\caption{First order estimation of power consumption in digital basic building blocks under accuracy scaling, based on simulations in a 40nm CMOS technology. The modelled blocks are multipliers ($\times$), adders ($+$), registers, wires and SRAM memory. In the power equation ($P$), only the switching activity $\alpha$ decreases as a function of the used number of bits $n$.  }
	\label{tab:powerConsumptionModels} 
	\centering
	\begin{array}{l|ccccc}
		P=\alpha C f V_{dd}^2& \times & $+$ & Reg. & Wire & SRAM\\
		\hline
		\alpha & O(n^2) & O(n) & O(n) & O(n) & O(n_{max}) \\
	\end{array}
	
\end{table}

Based on simulations in a 40nm technology, an energy model for common digital building blocks is proposed in Table~\ref{tab:powerConsumptionModels}. The reduced switching activity, and hence energy consumption, solely depends on the circuit architecture and can reduce quadratically (multiplier), or linearly (adders, register files, wiring).  
Using the energy-modelling from Table~\ref{tab:powerConsumptionModels}, we can estimate the impact of increased quantization arithmetic on the energy consumption of a full ConvNet-algorithms. 

In order to apply precision scaling efficiently in such a system, architectural changes have to be made. First, the positioning of the sign-bit in the fixed number representation is crucial. The MSB-bit should always be placed at the MSB position, otherwise toggling sign bits will lead to high switching activity. Second, The number of bits in a fixed point implementation tends to expand due to numerical operations. In order to make sure the number of bits remains limited throughout different stages of the algorithm, repetitive rounding or truncation is needed after every stage. Third, data- and parameter rescaling should be implemented between the convolutional layers using arithmetic shifting.
\begin{figure*}[!t]
		\centering
		\subfloat[]{{\label{fig:precisionScalingImpactOnZeroes_a}\includegraphics[width=0.45\textwidth]{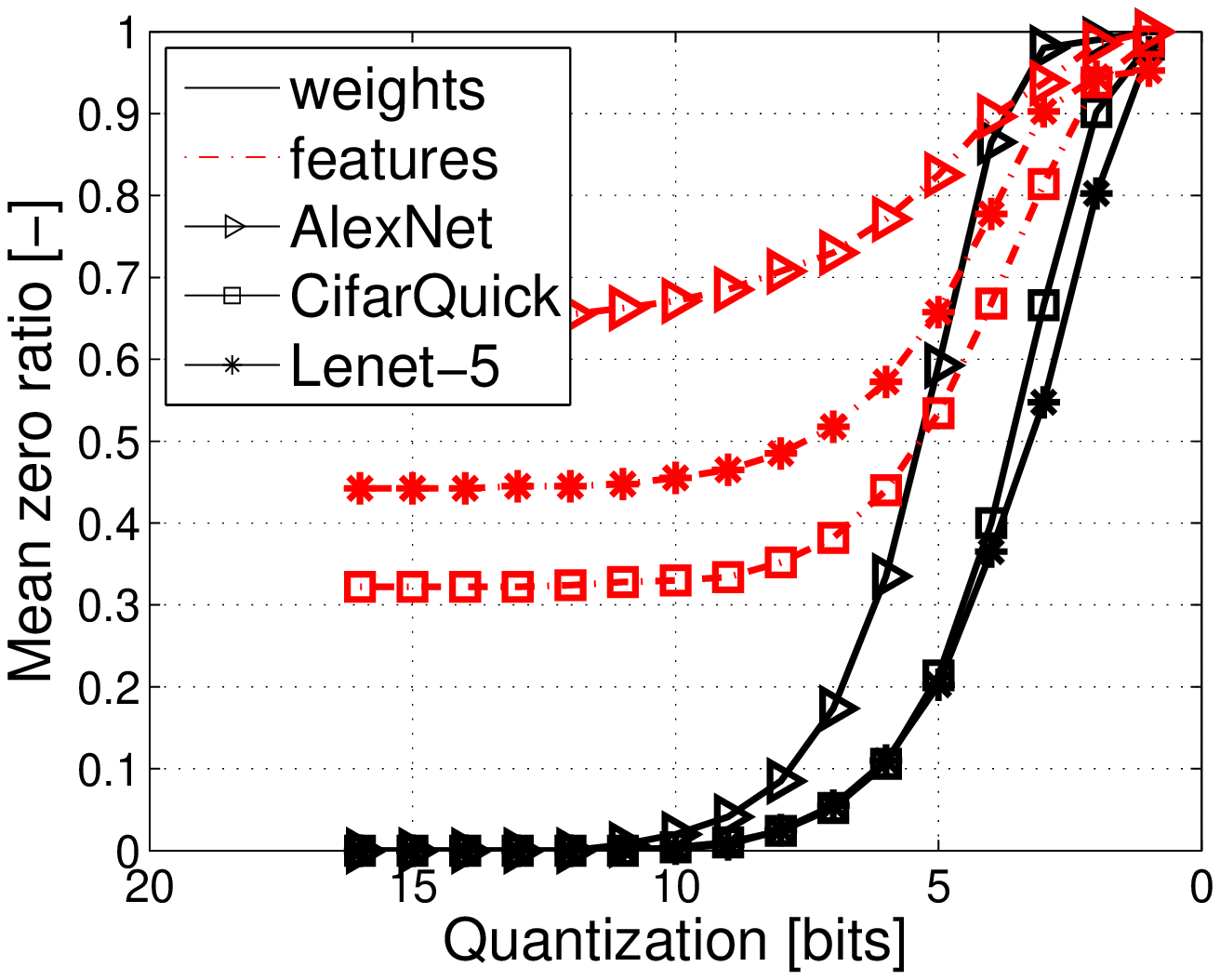} }}%
		\quad
		\subfloat[]{{\label{fig:precisionScalingImpactOnZeroes_b}\includegraphics[width=0.45\textwidth]{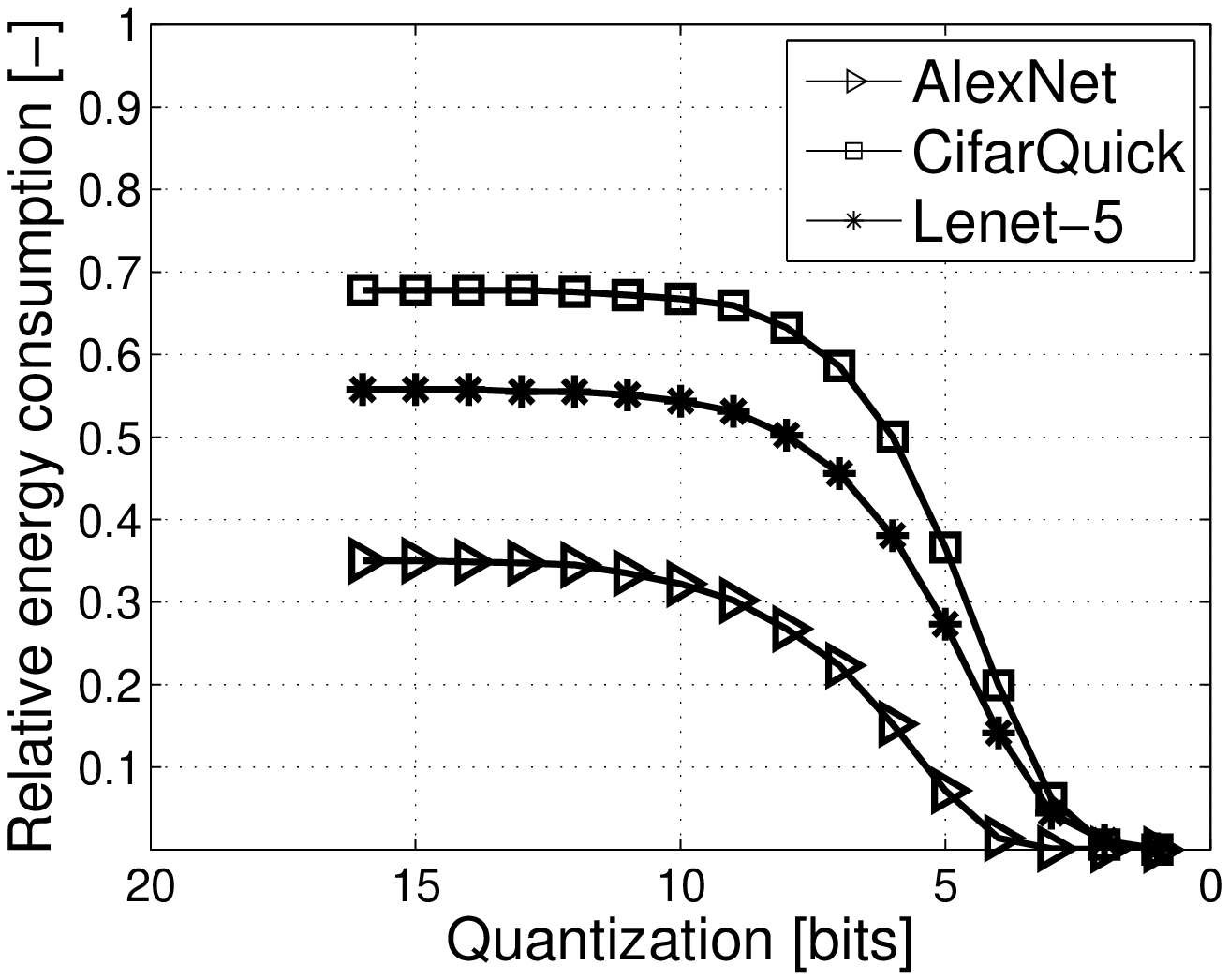} }}%
	\caption{(a) Impact of precision scaling on the average number of zeroes in the weights and input values of all three examples. (b) Relative energy reduction compared to the case without computation skipping.}%
	\label{fig:precisionScalingImpactOnZeroes}%
\end{figure*}

\subsubsection{Computation skipping}
An interesting feature of many modern convolutional neural networks is the appearance of Rectified Linear Unit layers (ReLU layers).
These put all negative inputs to zero and pass on positive values unchanged, as in $Output = max(0,Input)$. 
Since many layers in ConvNet-classification algorithms only output positive values when certain features are present, a large amount of ReLU-outputs will be zero and do not have to be used for further computations.
The ReLU layers thus allow for additional energy reductions by not computing unnecessary computations through computation skipping.

Figure~\ref{fig:precisionScalingImpactOnZeroes} shows the impact of precision scaling on the average number of zeroes in our ConvNet examples. For all architectures, the ratio of zeroed values lies between $50-90 \%$ of the total input values, depending on the used quantization setting. The number of zeroed values is significantly larger under precision scaling, from an average of $45\%$ on all LeNet-5 outputs at 16-bit to $65\%$ at 5-bit. 

It is very difficult to achieve such computation skipping in a software solution, since checking for zero-valued inputs in a scalar core consumes time. Using techniques for sparse matrix multiplication is not possible either, because the used convolutional kernels are typically very small on the order of $11\times11$, $5\times5$ or $3\times3$. 
However, computation skipping can be achieved in a hardware accelerator with dedicated hardware support. Flags can indicate if upcoming data is zero and prevent circuitry from switching if this is the case. As ConvNet-weights are fixed, these flags can be computed beforehand.   

\subsection{Achievable energy-accuracy trade-off in ConvNets}
\label{subsec:energyAccuracyTradeoff}

\begin{figure*}[!t]
		\centering
		\subfloat[]{{\label{fig:energy_vs_accuracy_a}\includegraphics[width=0.45\textwidth]{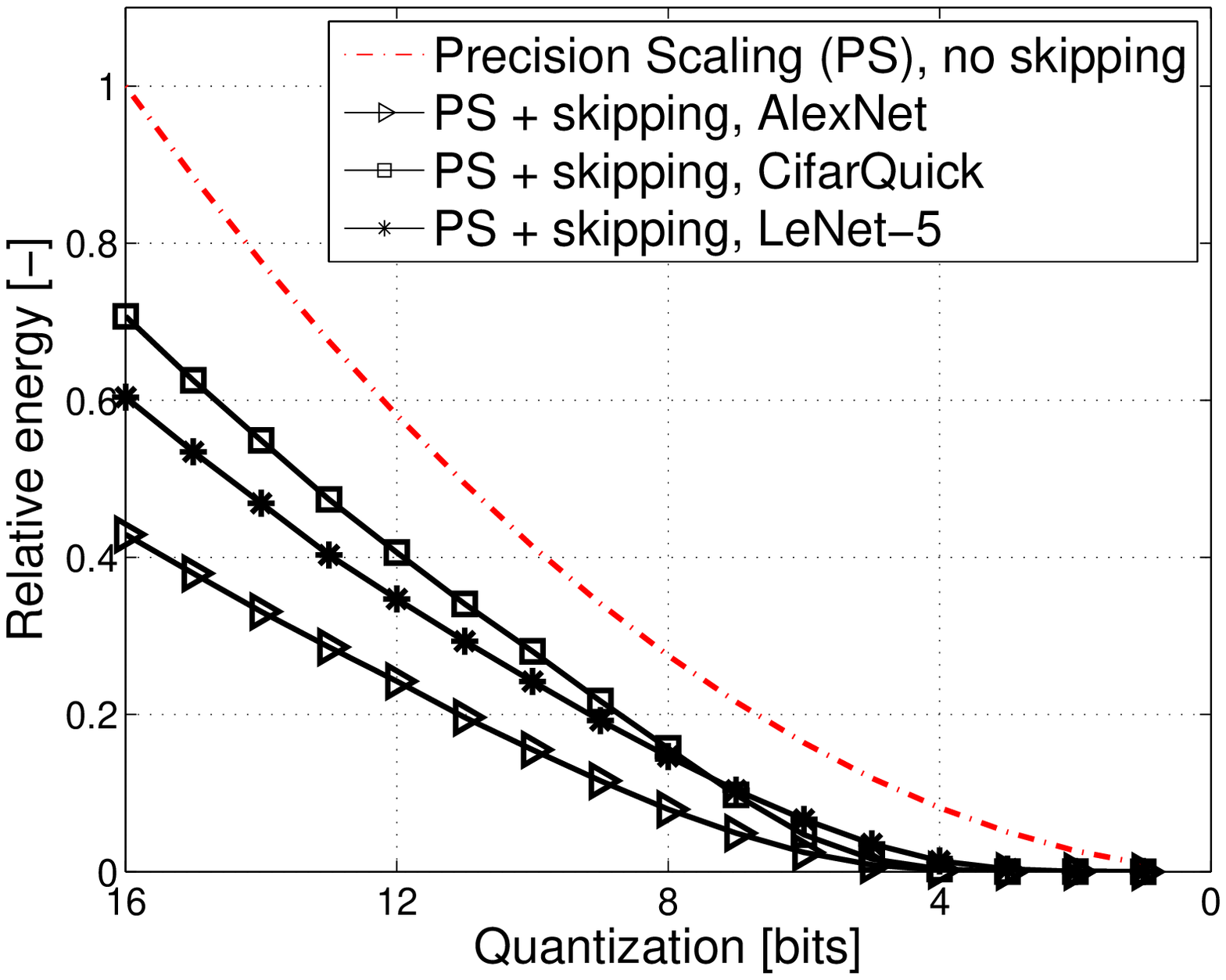} }}%
		\quad
		\subfloat[]{{\label{fig:energy_vs_accuracy_b}\includegraphics[width=0.45\textwidth]{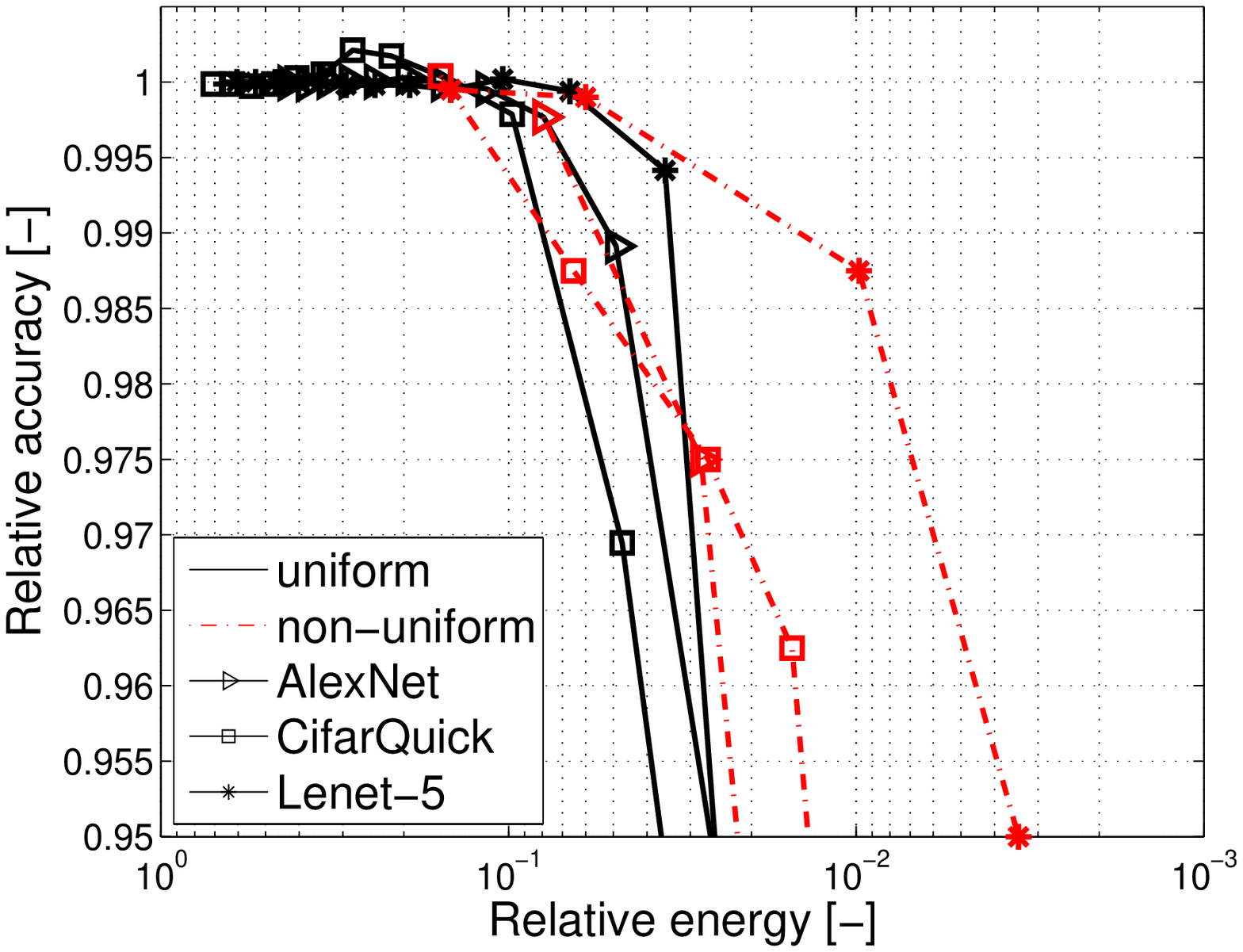} }}%
	\caption{ (a) Relative energy consumption for uniform quantization. A striking $5-20\times$ lower energy consumption at 8-bit, compared to the commonly used 16-bit used in previous literature.  (b) Energy-accuracy plot for per-layer quantization and scaling and with computation skipping. All considered ConvNet-networks gain at least an additional factor of $2$ in energy consumption if $99\%$ accuracy is allowed.}%
	\label{fig:energy_vs_accuracy}%
\end{figure*}

By combining both precision scaling and algorithmical computation skipping, major energy savings can be achieved in ConvNets. Figure~\ref{fig:energy_vs_accuracy_a} shows the effect of uniform quantization on the energy consumption of our benchmark algorithms, if both precision scaling and computation skipping are combined. Note how the energy reduces $5-20\times$ for an 8-bit implementation. 

Figure~\ref{fig:energy_vs_accuracy_b} shows curves in the energy-accuracy space for our ConvNet benchmarks.  It shows the trade-off for the $95-100\%$ classification accuracy window. All curves are given for both uniform and per-layer quantization, and are compared to the typically used uniform 16-bit number representation, without adequate precision scaling and computation skipping. If per-layer quantization is used, the trade-off is less steep than in the uniform case. This means more energy can be gained, while losing less classification accuracy in the process. All discussed networks gain at least an additional factor of 2 in energy consumption if a reduced classification accuracy of $99\%$ is allowed.

\begin{figure*}[!t]
	\begin{center}
		\centerline{
			\subfloat[]{{\label{fig:energyGains_a}\includegraphics[width=0.37\textwidth]{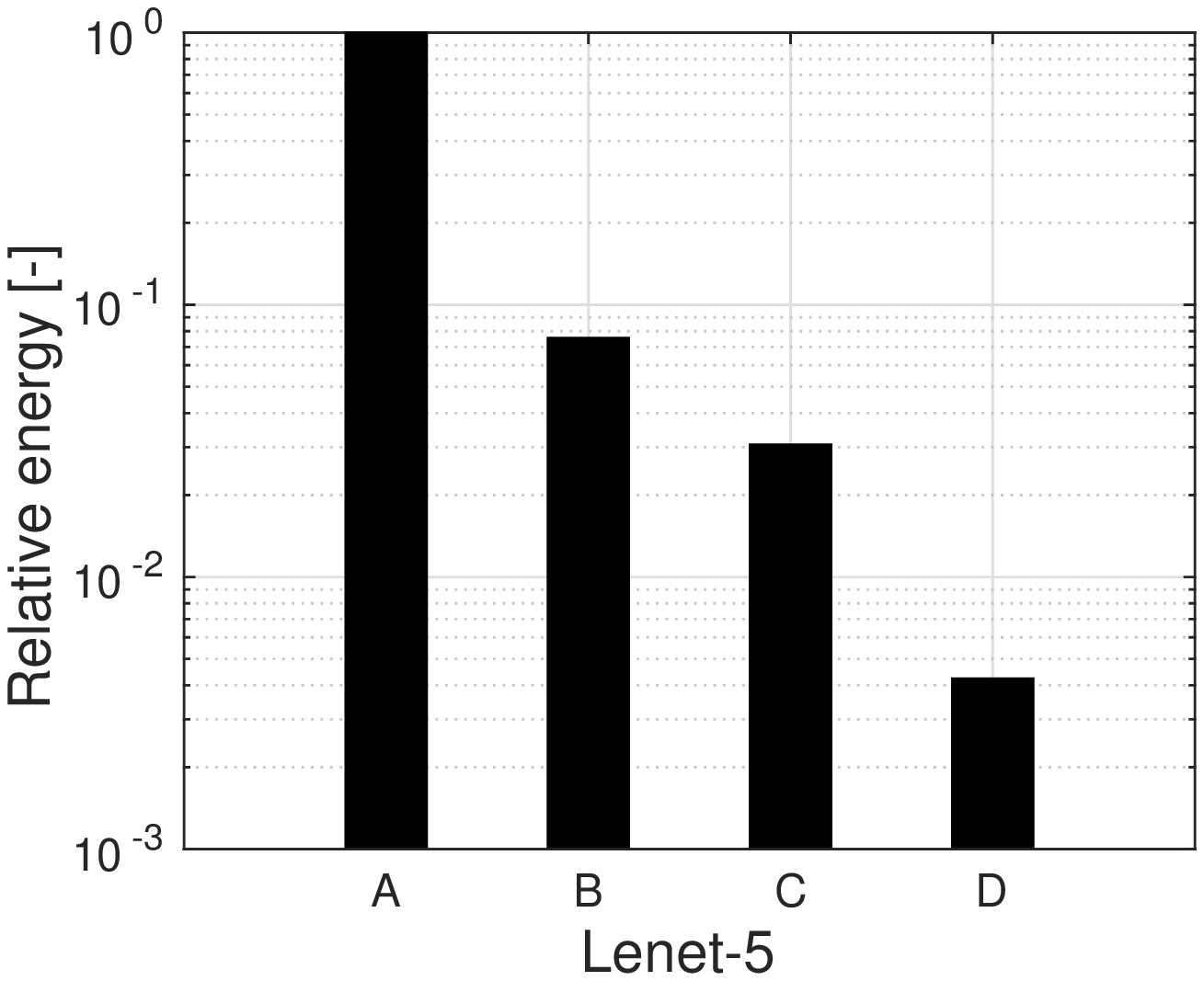} }}
			\subfloat[]{{\label{fig:energyGains_b}\includegraphics[width=0.37\textwidth]{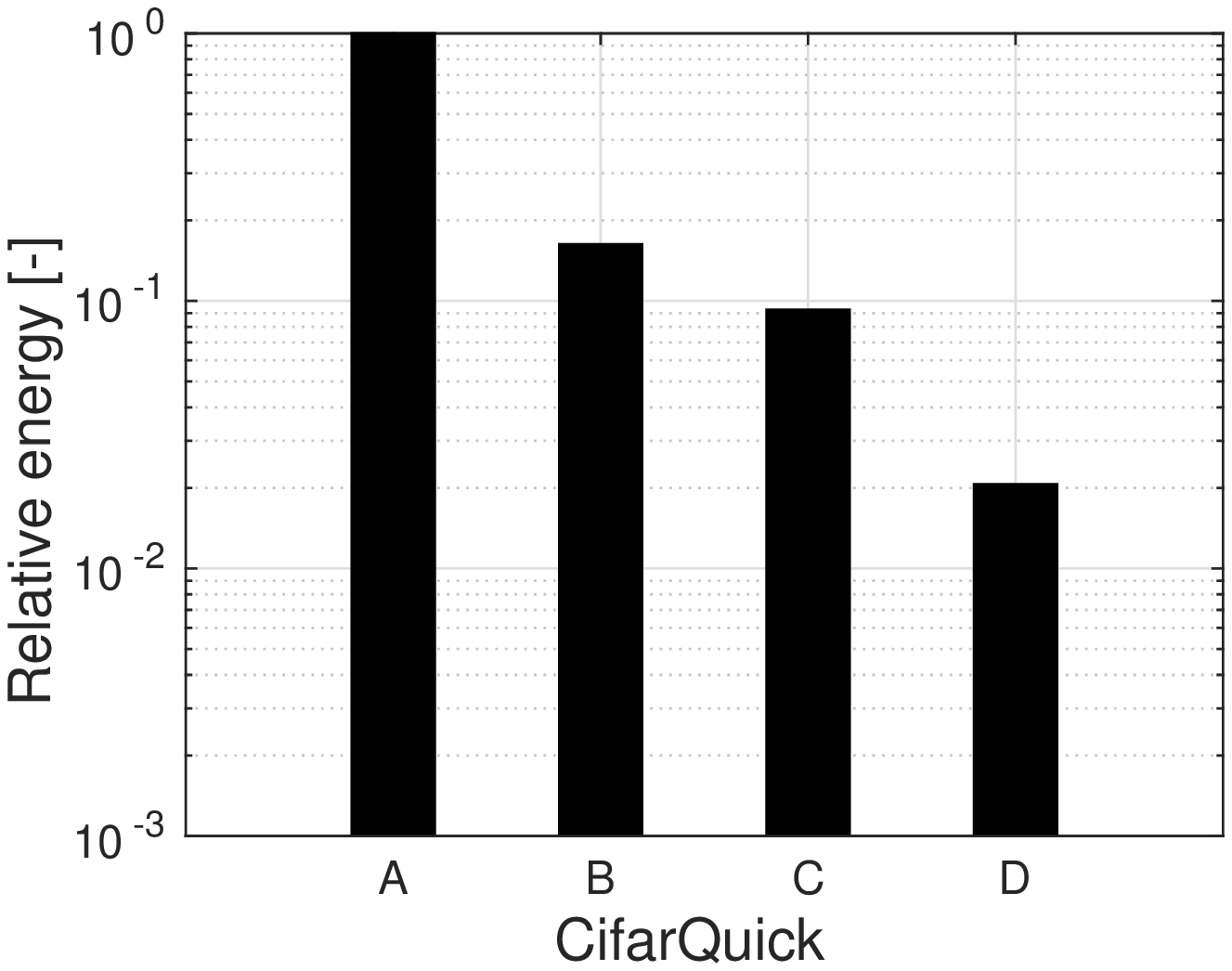} }}
			\subfloat[]{{\label{fig:energyGains_c}\includegraphics[width=0.37\textwidth]{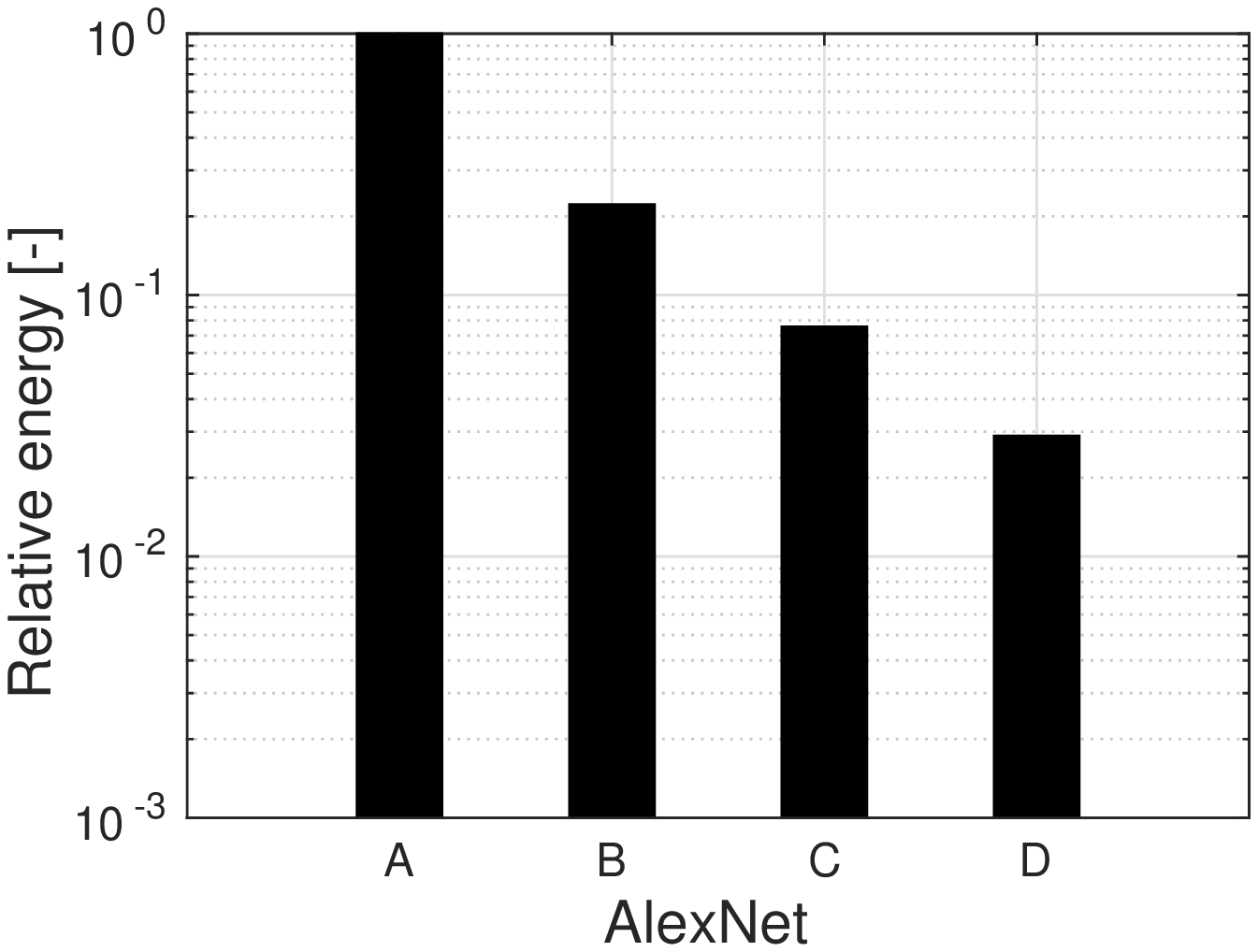} }}
		}
	\end{center}
	\caption{Energy comparison between our three benchmarks in different cases. (A) 16-bit (B) Per-layer scaling, uniform quantization at $100\%$ accuracy (C) Same as B, with computation skipping  (D) Per-layer scaling, per-layer quantization and computation skipping  at $99\%$ accuracy.}%
	\label{fig:energyGains}%
\end{figure*}

Figure~\ref{fig:energyGains} gives an overview of the effect of each improvement on the relative energy consumption. Per-layer rescaling and computation skipping lead to the largest energy gains.
Per-layer quantization leads to additional major energy gains if a reduced accuracy of $99\%$ can be allowed. 

\section{Conclusion}
ConvNet algorithms are typically very computation and memory intensive. 
In order to be able to embed ConvNet-based classification into wearable platforms, their energy consumption should be reduced significantly. 
In this work, we show how the energy consumption of dedicated ConvNet-accelerators can be drastically reduced by applying approximate computing. ConvNets for image classification are fault-tolerant and allow computation using 4-10 bits in stead of the previously used 16- or 32-bit formats. 
By using requantized number representations, the energy consumption in a ConvNet-accelerator can be reduced in two complementary ways. First, energy consumption can be lower through precision scaling. 
Second, more quantization further increases the number of zeroed parameter and input values in a ConvNet-algorithm. Since zero-valued numbers do not contribute to the algorithm's output, their computations can be skipped. 
The combination of these two techniques can lead to significant energy savings in several ConvNet-networks. We show energy reductions of up to $30\times$ compared to the commonly used 16-bit fixed point implementations, without sacrificing algorithm performance. If the classification accuracy can be reduced to $99\%$, additional energy savings can be achieved on the same network architecture, through per-layer optimization. In this case an additional $7.5\times$ reduction can be achieved.

{\small
\bibliographystyle{ieee}
\bibliography{bib/references}

\begin{thebibliography}{10}\itemsep=-1pt

\bibitem{chen2014diannao}
T.~Chen, Z.~Du, N.~Sun, J.~Wang, C.~Wu, Y.~Chen, and O.~Temam.
\newblock Diannao: A small-footprint high-throughput accelerator for ubiquitous
  machine-learning.
\newblock In {\em Proceedings of the 19th international conference on
  Architectural support for programming languages and operating systems}, pages
  269--284. ACM, 2014.

\bibitem{conti2015ultra}
F.~Conti and L.~Benini.
\newblock A ultra-low-energy convolution engine for fast brain-inspired vision
  in multicore clusters.
\newblock In {\em Proceedings of the 2015 Design, Automation \& Test in Europe
  Conference \& Exhibition}, pages 683--688. EDA Consortium, 2015.

\bibitem{de2012flexible}
M.~de~la Guia~Solaz, W.~Han, and R.~Conway.
\newblock A flexible low power dsp with a programmable truncated multiplier.
\newblock {\em Circuits and Systems I: Regular Papers, IEEE Transactions on},
  59(11):2555--2568, 2012.

\bibitem{deng_imagenet:_2009}
J.~Deng, W.~Dong, R.~Socher, L.-J. Li, K.~Li, and L.~Fei-Fei.
\newblock {ImageNet}: {A} large-scale hierarchical image database.
\newblock In {\em {IEEE} {Conference} on {Computer} {Vision} and {Pattern}
  {Recognition}, 2009. {CVPR} 2009}, pages 248--255, June 2009.

\bibitem{dundar1994effects}
G.~Dundar and K.~Rose.
\newblock The effects of quantization on multilayer neural networks.
\newblock {\em IEEE transactions on neural networks/a publication of the IEEE
  Neural Networks Council}, 6(6):1446--1451, 1994.

\bibitem{farabet2011neuflow}
C.~Farabet, B.~Martini, B.~Corda, P.~Akselrod, E.~Culurciello, and Y.~LeCun.
\newblock Neuflow: A runtime reconfigurable dataflow processor for vision.
\newblock In {\em Computer Vision and Pattern Recognition Workshops (CVPRW),
  2011 IEEE Computer Society Conference on}, pages 109--116. IEEE, 2011.

\bibitem{Jia2014}
Y.~Jia, E.~Shelhamer, J.~Donahue, S.~Karayev, J.~Long, R.~Girshick,
  S.~Guadarrama, and T.~Darrell.
\newblock Caffe: {Convolutional} {Architecture} for {Fast} {Feature}
  {Embedding}.
\newblock {\em arXiv:1408.5093 [cs]}, June 2014.
\newblock arXiv: 1408.5093.

\bibitem{jiang_effects_2003}
M.~Jiang and G.~Gielen.
\newblock The effects of quantization on multi-layer feedforward neural
  networks.
\newblock {\em International Journal of Pattern Recognition and Artificial
  Intelligence}, 17(04):637--661, June 2003.

\bibitem{krizhevsky2009learning}
A.~Krizhevsky.
\newblock Learning multiple layers of features from tiny images.
\newblock 2009.

\bibitem{krizhevsky_imagenet_2012}
A.~Krizhevsky, I.~Sutskever, and G.~E. Hinton.
\newblock {ImageNet} {Classification} with {Deep} {Convolutional} {Neural}
  {Networks}.
\newblock In F.~Pereira, C.~J.~C. Burges, L.~Bottou, and K.~Q. Weinberger,
  editors, {\em Advances in {Neural} {Information} {Processing} {Systems} 25},
  pages 1097--1105. Curran Associates, Inc., 2012.

\bibitem{kulkarni2011trading}
P.~Kulkarni, P.~Gupta, and M.~Ercegovac.
\newblock Trading accuracy for power with an underdesigned multiplier
  architecture.
\newblock In {\em VLSI Design (VLSI Design), 2011 24th International Conference
  on}, pages 346--351. IEEE, 2011.

\bibitem{le1990handwritten}
B.~B. Le~Cun, J.~S. Denker, D.~Henderson, R.~E. Howard, W.~Hubbard, and L.~D.
  Jackel.
\newblock Handwritten digit recognition with a back-propagation network.
\newblock In {\em Advances in neural information processing systems}. Citeseer,
  1990.

\bibitem{lecun1998mnist}
Y.~LeCun and C.~Cortes.
\newblock The mnist database of handwritten digits, 1998.

\bibitem{moons2015DVAS}
B.~Moons and M.~Verhelst.
\newblock Dvas: Dynamic voltage accuracy scaling for increased
  energy-efficiency in approximate computing.
\newblock {\em International Symposium on Low Power Electronics and Design
  (ISLPED)}, 2015.

\bibitem{peemen2013memory}
M.~Peemen, A.~A. Setio, B.~Mesman, and H.~Corporaal.
\newblock Memory-centric accelerator design for convolutional neural networks.
\newblock In {\em Computer Design (ICCD), 2013 IEEE 31st International
  Conference on}, pages 13--19. IEEE, 2013.

\bibitem{Simonyan14c}
K.~Simonyan and A.~Zisserman.
\newblock Very deep convolutional networks for large-scale image recognition.
\newblock {\em CoRR}, abs/1409.1556, 2014.

\bibitem{szegedy_going_2014}
C.~Szegedy, W.~Liu, Y.~Jia, P.~Sermanet, S.~Reed, D.~Anguelov, D.~Erhan,
  V.~Vanhoucke, and A.~Rabinovich.
\newblock Going {Deeper} with {Convolutions}.
\newblock {\em arXiv:1409.4842 [cs]}, Sept. 2014.
\newblock arXiv: 1409.4842.

\bibitem{venkataramani2013quality}
S.~Venkataramani, V.~K. Chippa, S.~T. Chakradhar, K.~Roy, and A.~Raghunathan.
\newblock Quality programmable vector processors for approximate computing.
\newblock In {\em Proceedings of the 46th Annual IEEE/ACM International
  Symposium on Microarchitecture}, pages 1--12. ACM, 2013.

\bibitem{venkataramani2014axnn}
S.~Venkataramani, A.~Ranjan, K.~Roy, and A.~Raghunathan.
\newblock Axnn: energy-efficient neuromorphic systems using approximate
  computing.
\newblock In {\em Proceedings of the 2014 international symposium on Low power
  electronics and design}, pages 27--32. ACM, 2014.

\bibitem{venkataramani2012salsa}
S.~Venkataramani, A.~Sabne, V.~Kozhikkottu, K.~Roy, and A.~Raghunathan.
\newblock Salsa: systematic logic synthesis of approximate circuits.
\newblock In {\em Proceedings of the 49th Annual Design Automation Conference},
  pages 796--801. ACM, 2012.

\bibitem{wu_deep_2015}
R.~Wu, S.~Yan, Y.~Shan, Q.~Dang, and G.~Sun.
\newblock Deep {Image}: {Scaling} up {Image} {Recognition}.
\newblock {\em arXiv:1501.02876 [cs]}, Jan. 2015.
\newblock arXiv: 1501.02876.

\end{thebibliography}
}

\end{document}